%% file: main.tex
\pdfoutput=1

\documentclass[11pt]{article}

\usepackage{emnlp2021}

\usepackage{times}
\usepackage{latexsym}

\usepackage[T1]{fontenc}

\usepackage[utf8]{inputenc}

\usepackage{microtype}

\input{content/myPackages}

\input{content/myNotations}

\newcommand{\mn}{SAECON}  

\newcommand{\myTitle}{Powering Comparative Classification with Sentiment Analysis via Domain Adaptive Knowledge Transfer}
\title{\myTitle}

\usepackage{lipsum}

\newcommand\blfootnote[1]{%
  \begingroup
  \renewcommand\thefootnote{}\footnote{#1}%
  \addtocounter{footnote}{-1}%
  \endgroup
}

\author{Zeyu Li, Yilong Qin*, Zihan Liu*, and Wei Wang\\
  Department of Compute Science, University of California, Los Angeles \\
  \texttt{\{zyli,weiwang\}@cs.ucla.edu} \\
  \texttt{\{louisqin,zihanliu\}@ucla.edu}
}

\begin{document}
\maketitle
\blfootnote{* These authors contributed equally.}
\begin{abstract}
\input{content/0abstract}
\end{abstract}

\input{content/1intro}


\input{content/2relatedwork}

\input{content/3method}

\input{content/4experiments}

\input{content/5conclusion}
\input{content/8acknowledgement}

\input{content/7broaderimpact}
\bibliography{references}
\bibliographystyle{acl_natbib}

\clearpage
\appendix

\input{content/6appendix}

\end{document}

%% file: content/myPackages.tex
\usepackage{url}
\usepackage{graphics}
\usepackage{amsmath}
\usepackage{amssymb}
\usepackage{amsfonts}
\usepackage{mathtools}
\usepackage{booktabs}
\usepackage{subcaption}
\usepackage{siunitx}
\usepackage{booktabs}
\usepackage{colortbl}
\usepackage{array}
\usepackage{soul}
\usepackage{pifont}
\usepackage{tabularx}
\usepackage{multirow}
\usepackage{tikz-dependency}
\usepackage[linesnumbered,ruled,vlined]{algorithm2e}
\usepackage{textcomp}
\usepackage{pifont}
\usepackage{xcolor}
\usepackage{bbm}

%% file: content/myNotations.tex
\newcommand{\matW}{\boldsymbol{\mathrm{W}}}

\newcommand{\matS}{\boldsymbol{\mathrm{S}}}

\newcommand{\matI}{\boldsymbol{\mathrm{I}}}

\newcommand{\vecx}{\boldsymbol{x}}
\newcommand{\vecw}{\boldsymbol{w}}

\newcommand{\vecb}{\boldsymbol{b}}
\newcommand{\vecd}{\boldsymbol{d}}

\newcommand{\vece}{\boldsymbol{e}}

\newcommand{\vech}{\boldsymbol{h}}

\newcommand{\vecbeta}{\boldsymbol{\beta}}
\newcommand{\veclambda}{\boldsymbol{\lambda}}

\DeclareMathOperator{\sm}{\delta}

\newcommand{\calF}{\mathcal{F}}

\newcommand{\xmark}{\ding{55}}  
\newcommand{\cmark}{\ding{51}}

%% file: content/0abstract.tex
We study Comparative Preference Classification (CPC) which aims at predicting whether a preference comparison exists between two entities in a given sentence and, if so, which entity is preferred over the other.
High-quality CPC models can significantly benefit applications such as comparative question answering and review-based recommendation.
Among the existing approaches, non-deep learning methods suffer from inferior performances.
The state-of-the-art graph neural network-based ED-GAT~\cite{bingliu20} only considers syntactic information while ignoring the critical semantic relations and the sentiments to the compared entities.
We propose Sentiment Analysis Enhanced COmparative Network (\mn{}) which improves CPC accuracy with a sentiment analyzer that learns sentiments to individual entities via domain adaptive knowledge transfer. Experiments on the CompSent-19~\cite{panchenko19} dataset present a significant improvement on the F1 scores over the best existing CPC approaches.

%% file: content/1intro.tex
\section{Introduction}
\label{sec:intro}







Comparative Preference Classification (CPC) is a natural language processing (NLP) task that predicts whether a preference comparison exists between two entities in a sentence and, if so, which entity wins the game. For example, given the sentence: \textit{Python is better suited for data analysis than MATLAB due to the many available deep learning libraries}, a decisive comparison exists between \textit{Python} and \textit{MATLAB} and comparatively \textit{Python} is preferred over \textit{MATLAB} in the context.

The CPC task can profoundly impact various real-world application scenarios. Search engine users may query not only factual questions but also comparative ones to meet their specific information needs~\cite{gupta17identifying}. Recommendation providers can analyze product reviews with comparative statements to understand the advantages and disadvantages of the product comparing with similar ones.

Several models have been proposed to solve this problem. \citet{panchenko19} first formalize the CPC problem, build and publish the CompSent-19 dataset, and experiment with numerous general machine learning models such as Support Vector Machine (SVM), representation-based classification, and XGBoost.
However, these attempts consider CPC as a sentence classification while ignoring the semantics and the contexts of the entities~\cite{bingliu20}.

ED-GAT~\cite{bingliu20} marks the first entity-aware CPC approach that captures long-distance \textit{syntactic} relations between the entities of interest by applying graph attention networks (GAT) to dependency parsing graphs. However, we argue that the disadvantages of such an approach are clear. Firstly, ED-GAT replaces the entity names with ``entityA'' and ``entityB'' for simplicity and hence deprives their \textit{semantics}. Secondly, ED-GAT has a deep architecture with ten stacking GAT layers to tackle the long-distance issue between compared entities. However, more GAT layers result in a heavier computational workload and reduced training stability. Thirdly, although the competing entities are typically connected via multiple hops of dependency relations, the unordered tokens along the connection path cannot capture either global or local high-quality semantic context features.

In this work, we propose a Sentiment Analysis Enhanced COmparative classification Network (\mn{}), a CPC approach that considers not only syntactic but also semantic features of the entities.
The semantic features here refer to the context of the entities from which a sentiment analysis model can infer the sentiments toward the entities. 
Specifically, the encoded sentence and entities are fed into a dual-channel context feature extractor to learn the global and local context. In addition, an auxiliary Aspect-Based Sentiment Analysis (ABSA) module is integrated to learn the \textit{sentiments} towards individual entities which are greatly beneficial to the comparison classification.

ABSA aims to detect the specific emotional inclination toward an aspect within a sentence~\cite{DBLP:conf/aaai/MaPC18,hu-etal-2019-open,phan-ogunbona-2020-modelling,absa1,relational2020}. For example, the sentence \textit{I liked the service and the staff but not the food} suggests positive sentiments toward service and staff but a negative one toward food. These \textit{aspect} entities, such as service, staff, and food, are studied individually. 

The well-studied ABSA approaches can be beneficial to CPC when the compared entities in a CPC sentence are considered as the aspects in ABSA. Incorporating the individual sentiments learned by ABSA methods into CPC has several advantages. Firstly, for a comparison to hold, the preferred entity usually receives a positive sentiment while its rival gets a relatively negative one. These sentiments can be easily extracted by the strong ABSA models. The contrast between the sentiments assigned to the compared entities provides a vital clue for an accurate CPC.
Secondly, the ABSA models are designed to target the sentiments toward phrases, which bypasses the complicated and noisy syntactic relation path. 
Thirdly, considering the scarcity of 
the data resource of CPC, the abundant annotated data of ABSA can provide sufficient supervision signal to improve the accuracy of CPC.

There is one challenge that blocks the knowledge transfer of sentiment analysis from the ABSA data to the CPC task: \textit{domain shift}. Existing ABSA datasets are centered around specific topics such as restaurants and laptops, while the CPC data has mixed topics~\cite{panchenko19} that are all distant from restaurants. 
In other words, sentences of ABSA and CPC datasets are drawn from different distributions, also known as \textit{domains}. The difference in the distributions is referred to as a ``domain shift''~\cite{ganin2015unsupervised,xdomNLP} and it is harmful to an accurate knowledge transfer.
To mitigate the domain shift, we design a domain adaptive layer to remove the domain-specific feature such as topics and preserve the domain-invariant feature such as sentiments of the text so that the sentiment analyzer can smoothly transfer knowledge from sentiment analysis to comparative classification. 
\input{figures/pipeline}

%% file: figures/pipeline.tex
\begin{figure*}[!h]
    \centering
    \includegraphics[width=0.95\linewidth]{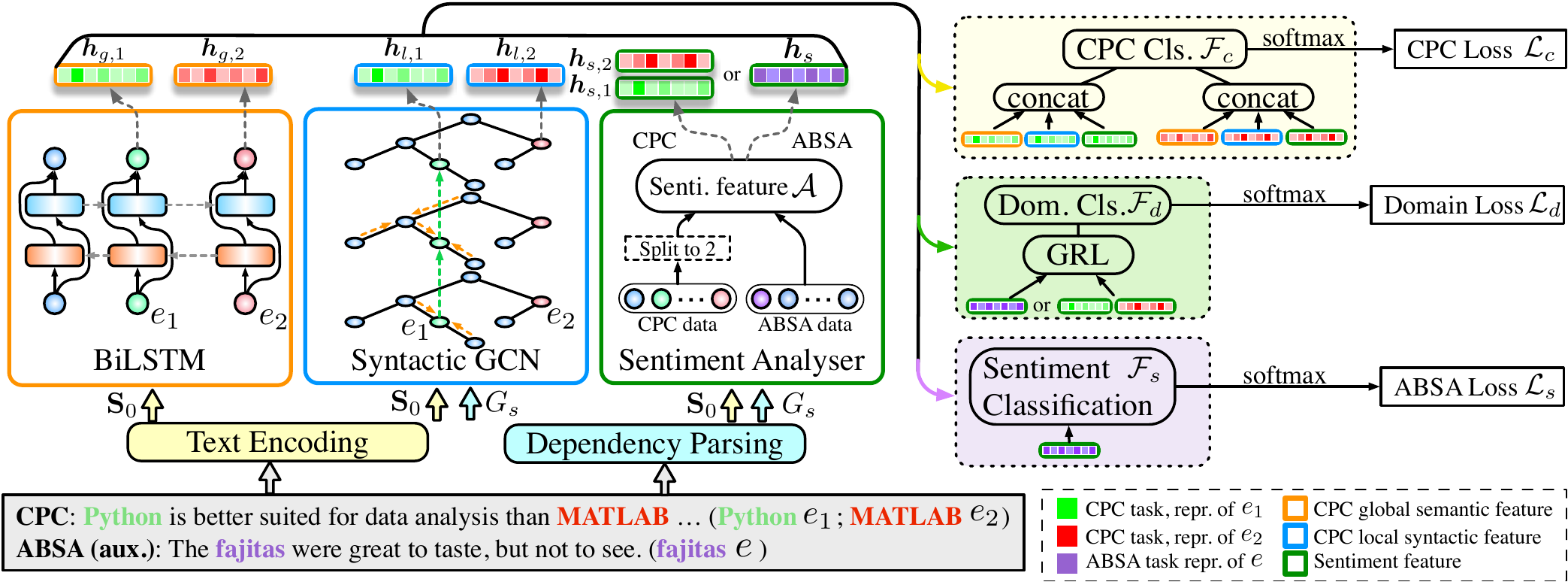}
    \caption{Pipeline of \mn{}. Sentences from two domains shown in the gray box are fed into text encoder and dependency parser. The resultant representations of the entities from three substructures are discriminated by different colors (see the legend in the corner). ``Cls.'' is short for classifier.}
    \label{fig:pipeline}
\end{figure*}

%% file: content/2relatedwork.tex
\section{Related Work}
\label{sec:related_work}
\subsection{Comparative Preference Classification}
CPC originates from the task of Comparative Sentence Identification (CSI)~\cite{DBLP:conf/sigir/JindalL06}. CSI aims to identify the comparative sentences. \citet{DBLP:conf/sigir/JindalL06} approach this problem by Class Sequential Mining (CSR) and a Naive Bayesian classifier.
Building upon CSI,~\citet{panchenko19} propose the task of CPC, release CompSent-19 dataset, and conduct experimental studies using traditional machine learning approaches such as SVM, representation-based classification, and XGBoost. However, they neglect the entities in the comparative context~\cite{panchenko19}.
ED-GAT~\cite{bingliu20}, a more recent work, uses the dependency graph to better recognize long-distance comparisons and avoid falsely identifying unrelated comparison predicates. However, it fails to capture semantic information of the entities as they are replaced by ``entityA'' and ``entityB''. Furthermore, having multiple GAT layers severely increases training difficulty.
\subsection{Aspect-Based Sentiment Analysis}
ABSA derives from sentiment analysis (SA) which infers the sentiment associated with a specific entity in a sentence. Traditional approaches of ABSA utilize SVM for classification~\cite{kiritchenko-etal-2014-nrc,wagner-etal-2014-dcu, zhang-etal-2014-ecnu} while neural network-based approaches employ variants of RNN~\cite{nguyen-shirai-2015-phrasernn,DBLP:journals/access/AydinG20},  LSTM~\cite{tang-etal-2016-effective,wang-etal-2016-attention,bao-etal-2019-attention}, GAT~\cite{relational2020}, and GCN~\cite{pouran-ben-veyseh-etal-2020-improving,DBLP:journals/access/XuZL20}.

More recent works\footnote{Due to the limited space, we are unable to exhaustively cover all references. Works discussed are classic examples.} widely use complex contextualized NLP models such as BERT~\cite{bert}. \citet{sun-etal-2019-utilizing} transform ABSA into a Question Answering task by constructing auxiliary sentences. \citet{phan-ogunbona-2020-modelling} build a pipeline of Aspect Extraction and ABSA and used wide and concentrated features for sentiment classification.
ABSA is related to CPC by nature. In general, entities with positive sentiments are preferred over the ones with neutral or negative sentiments. Therefore, the performance of a CPC model can be enhanced by the ABSA techniques.


%% file: content/3method.tex
\section{\mn{}}
In this section, we first formalize the problem and then explain \mn{} in detail.
The pipeline of \mn{} is depicted in Figure~\ref{fig:pipeline} with essential notations. 
\subsection{Problem Statement}
\paragraph{CPC} 
Given a sentence $s$
from the CPC corpus $D_c$ with $n$ tokens and two entities $e_1$ and $e_2$, a CPC model predicts whether there exists a preference comparison between $e_1$ and $e_2$ in $s$ and if so, which entity is preferred over the other. Potential results can be \texttt{Better} ($e_1$  wins), \texttt{Worse} ($e_2$ wins), or \texttt{None} (no comparison exists). 

\paragraph{ABSA} 
Given a sentence $s'$
from the ABSA corpus $D_s$ with $m$ tokens and one entity $e'$, ABSA identifies the sentiment (positive, negative, or neutral) associated with $e'$. 

We denote the source domains of the CPC and ABSA datasets by $\mathcal{D}_c$ and $\mathcal{D}_s$. $D_c$ and $D_s$ contain samples that are drawn from $\mathcal{D}_c$ and $\mathcal{D}_s$, respectively. $\mathcal{D}_c$ and $\mathcal{D}_s$ are similar but different in topics which produces a domain shift.
We use $s$ to denote sentences in $D_c \cup D_s$ and $E$ to denote the entity sets for simplicity in later discussion. $|E|=2$ if $s\in D_c$ and $|E|=1$ otherwise.

\subsection{Text Feature Representations}
A sentence is encoded by its word representations via a text encoder and parsed into a dependency graph via a dependency parser~\cite{chen2014fast}. 
Text encoder, such as GloVe~\cite{pennington2014glove} and BERT~\cite{bert}, maps a word $w$ into a low dimensional embedding $\vecw\in\mathbb{R}^{d_0}$. 
GloVe assigns a fixed vector while BERT computes a token\footnote{BERT generates representations of wordpieces which can be substrings of words. If a word is broken into wordpieces by BERT tokenizer, the average of the wordpiece representations is taken as the word representation. The representations of the special tokens of BERT, \texttt{[CLS]} and \texttt{[SEP]}, are not used.} representation by its textual context. The encoding output of $s$ is denoted by 
$\matS_0= \{\vecw_1,\ldots, \vece_1, \dots, \vece_2, \dots, \vecw_n\}$ where $\vece_i$ denotes the embedding of entity $i$, $\vecw_i$ denotes the embedding of a non-entity word, and $\vecw_i,\vece_j\in\mathbb{R}^{d_0}$.

The dependency graph of $s$, denoted by $G_s$, is obtained by applying a dependency parser to $s$ such as Stanford Parser~\cite{chen2014fast} or spaCy\footnote{\url{https://spacy.io}}.
$G_s$ is a syntactic view of $s$~\cite{titov2017,ggnn} that is composed of vertices of words and directed edges of dependency relations. 
Advantageously, complex syntactic relations between distant words in the sentence can be easily detected with a small number of hops over dependency edges~\cite{bingliu20}. 

\subsection{Contextual Features for CPC}
\label{subsec:cpc_features}
\paragraph{Global Semantic Context}
To model more extended context of the entities, we use a bi-directional LSTM (BiLSTM) to encode the entire sentence in both directions. 
Bi-directional recurrent neural network is widely used in extracting semantics~\cite{li2019personalized}.
Given the indices of $e_1$ and $e_2$ in $s$, the global context representations $\vech_{g,1}$ and $\vech_{g,2}$ are computed by averaging the hidden outputs from both directions.
\begin{align*}
\overrightarrow{\vech_{g,i}}, \overleftarrow{\vech_{g,i}} &= \text{BiLSTM}(\matS_0)[e_i.\text{index}], \quad i=1, 2 \\
\vech_{g,i} &= \frac{1}{2}\left(\overrightarrow{\vech_{g,i}}+ \overleftarrow{\vech_{g,i}}\right), \vech_{g,i}\in \mathbb{R}^{d_g}.
\end{align*}

\paragraph{Local Syntactic Context}
In \mn{}, we use a dependency graph to capture the syntactically neighboring context of entities that contains words or phrases modifying the entities and indicates comparative preferences. We apply a Syntactic Graph Convolutional Network (SGCN)~\cite{sgcn,titov2017} to $G_s$ to compute the local context feature $\vech_{l,1}$ and $\vech_{l,2}$ for $e_1$ and $e_2$, respectively. SGCN operates on directed dependency graphs with three major adjustments compared with GCN~\cite{gcn}: considering the directionality of edges, separating parameters for different dependency labels\footnote{Labels are defined as the combinations of directions and dependency types. For example, edge $((u,v),$ \texttt{nsubj}$)$ and edge $((v,u),$ \texttt{nsubj}$^{-1})$ have different labels.}, and applying edge-wise gating to message passing.

GCN is a multilayer message propagation-based graph neural network. Given a vertex $v$ in $G_s$ and its neighbors $\mathcal{N}(v)$, the vertex representation of $v$ on the $(j+1)$th layer is given as
\begin{equation*}
\vech_v^{(j+1)} = \rho \left(\sum_{u\in\mathcal{N}(v)}\matW^{(j)}\vech_u^{(j)}+\vecb^{(j)}\right),
\end{equation*}
where $\rho(\cdot)$ denotes an aggregation function such as mean and sum, $\matW^{(j)}\in \mathbb{R}^{d^{(j+1)}\times d^{(j)}}$ and $\vecb^{(j)}\in\mathbb{R}^{d^{(j+1)}}$ are trainable parameters, and $d^{(j+1)}$ and $d^{(j)}$ denote latent feature dimensions of the $(j+1)$th and the $j$th layers, respectively.

SGCN improves GCN by considering different edge directions and diverse edge types, and assigns different parameters to different directions or labels. 
However, there is one caveat: the directionality-based method cannot accommodate the rich edge type information; the label-based method causes combinatorial over-parameterization, increased risk of overfitting, and reduced efficiency. Therefore, we naturally arrive at a trade-off of using direction-specific weights and label-specific biases.

The edge-wise gating can select impactful neighbors by controlling the gates for message propagation through edges. The gate on the $j$th layer of an edge between vertices $u$ and $v$ is defined as 
$$g^{(j)}_{uv}=\sigma\left(\vech_u^{(j)}\cdot\vecbeta^{(j)}_{d_{uv}}+\gamma^{(j)}_{l_{uv}}\right),\quad g^{(j)}_{uv}\in\mathbb{R},$$
where $d_{uv}$ and $l_{uv}$ denote the direction and label of edge $(u,v)$, $\vecbeta^{(j)}_{d_{uv}}$ and $\gamma^{(j)}_{l_{uv}}$ are trainable parameters, and $\sigma(\cdot)$ denotes the sigmoid function.

Summing up the aforementioned adjustments on GCN, the final vertex representation learning is 
\begin{equation*}
\vech_v^{(j+1)} = \rho \left(\sum_{u\in\mathcal{N}(v)}g^{(j)}_{uv}\left(\matW^{(j)}_{d_{uv}}\vech_u^{(j)}+\vecb_{l_{uv}}^{(j)}\right)\right).
\end{equation*}
Vectors of $\matS_0$ serve as the input representations $\vech^{(0)}_v$ to the first SGCN layer. The representations corresponding to $e_1$ and $e_2$ are the output $\{\vech_{l,1}, \vech_{l,2}\}$ with dimension $d_l$.

\subsection{Sentiment Analysis with Knowledge Transfer from ABSA}
\label{subsec:knowledge_transfer}
We have discussed in Section~\ref{sec:intro} that ABSA inherently correlates with the CPC task. Therefore, it is natural to incorporate a sentiment analyzer into \mn{} as an auxiliary task to take advantage of the abundant training resources of ABSA to boost the performance on CPC.
There are two paradigms for auxiliary tasks: (1) incorporating fixed parameters that are pretrained solely with the auxiliary dataset; 
(2) incorporating the architecture only with untrained parameters and jointly optimizing them from scratch with the main task simultaneously~\cite{Li_Wei_Zhang_Yang_2018,xdomNLP,wang2018recursive}.

Option (1) ignores the domain shift between $D_c$ and $D_s$, which degrades the quality of the learned sentiment features since the domain identity information is noisy and unrelated to the CPC task. \mn{} uses option (2).
For a smooth and efficient knowledge transfer from $\mathcal{D}_s$ to $\mathcal{D}_c$ under the setting of option (2), the ideal sentiment analyzer only extracts the textual feature that is contingent on sentimental information but orthogonal to the identity of the source domain.
In other words, the learned sentiment features are expected to be \textit{discriminative} on sentiment analysis but \textit{invariant} with respect to the domain shift. Therefore, the sentiment features are more aligned with the CPC domain $\mathcal{D}_c$ with reduced noise from domain shift.

In \mn{}, we use a \textit{gradient reversal layer} (GRL) and a \textit{domain classifier} (DC)~\cite{ganin2015unsupervised} for the domain adaptive sentiment feature learning that maintains the discriminativeness and the domain-invariance. 
GRL+DC is a straightforward, generic, and effective modification to neural networks for domain adaptation~\cite{kamath-etal-2019-reversing,gu-etal-2019-improving,belinkov-etal-2019-adversarial,Li_Wei_Zhang_Yang_2018}. It can effectively close the shift between complex distributions~\cite{ganin2015unsupervised} such as $\mathcal{D}_c$ and $\mathcal{D}_s$.

Let $\mathcal{A}$ denote the sentiment analyzer which alternatively learns sentiment information from $D_s$ and provides sentimental clues to the compared entities in $D_c$. Specifically, each CPC instance is split into two ABSA samples with the same text before being fed into $\mathcal{A}$ (see the ``Split to 2'' in Figure~\ref{fig:pipeline}). One takes $e_1$ as the queried aspect the other takes $e_2$.
\begin{equation*}
\mathcal{A}(\matS_0, G_s, E)=\left\{
    \begin{aligned}
        &\vech_{s,1}, \vech_{s,2} &\text{ if }s\in D_c,\\
        &\vech_{s}        &\text{ if }s\in D_s.\\
    \end{aligned}
\right.
\end{equation*}
$\vech_{s,1}$, $\vech_{s,2}$, and $\vech_{s} \in \mathbb{R}^{d_s}$.
These outputs are later sent through a GRL to not only the CPC and ABSA predictors shown in Figure~\ref{fig:pipeline} but also the DC to predict the source domain $y_d$ of $s$ where $y_d=1$ if $s\in D_s$ otherwise $0$. GRL, trainable by backpropagation, is transparent in the forward pass ($\text{GRL}_\alpha(\vecx) = \vecx$). It reverses the gradients in the backward pass as $$\frac{\partial \text{GRL}_\alpha}{\partial \vecx} = -\alpha \matI.$$
Here $\vecx$ is the input to GRL, $\alpha$ is a hyperparameter, and $\matI$ is an identity matrix.
During training, the reversed gradients maximize the domain loss, forcing $\mathcal{A}$ to forget the domain identity via the backpropagation and mitigating the domain shift.
Therefore, the outputs of $\mathcal{A}$ stay invariant to the domain shift.
But as the outputs of $\mathcal{A}$ are also optimized for ABSA predictions, the distinctiveness with respect to sentiment classification is retained. 

Finally, the selection of $\mathcal{A}$ is flexible as it is architecture-agnostic. In this paper, we use the LCF-ASC aspect-based sentiment analyzer proposed by~\citet{phan-ogunbona-2020-modelling} in which two scales of representations are concatenated to learn the sentiments to the entities of interest.

\subsection{Objective and Optimization}
\label{subsec:obj_optim}
\mn{} optimizes three classification errors overall for CPC, ABSA, and domain classification. 
For CPC task, features for local context, global context, and sentiment are concatenated: $\vech_{e_i} = [\vech_{g,i}; \vech_{l,i}; \vech_{s,i}]$, $i\in\{1,2\}$, and $\vech_{e_i}\in\mathbb{R}^{d_s+d_g+d_l}$.
Given $\calF_c$, $\calF_s$, $\calF_d$, and $\calF$ below denoting fully-connected neural networks with non-linear activation layers,
CPC, ABSA, domain predictions are obtained by
\begin{align*}
    \hat{y}_c   &= \sm(\calF_c([\calF(\vech_{e_1});\calF(\vech_{e_2})])) &\text{ (CPC only),}\\
    \hat{y}_s   &= \sm(\calF_s(\vech_s)) &\text{ (ABSA only),}  \\
    \hat{y}_d   &= \sm(\calF_d(\text{GRL}(\mathcal{A}(\matS_0, G_s, E)))) &\text{ (Both tasks),}
\end{align*}
where $\delta$ denotes the softmax function. With the predictions, \mn{} computes the cross entropy losses for the three tasks as $\mathcal{L}_c$, $\mathcal{L}_s$, and $\mathcal{L}_d$, respectively. The label of $\mathcal{L}_d$ is $y_d$. The computations of the losses are omitted due to the space limit.

In summary, the objective function of the proposed model \mn{} is given as follows,
\begin{equation*}
    \mathcal{L} = \mathcal{L}_c + \lambda_s \mathcal{L}_s + \lambda_d \mathcal{L}_d + \lambda \text{reg}(L2),
\end{equation*}
where $\lambda_s$ and $\lambda_d$ are two weights of the losses, and $\lambda$ is the weight of an $L2$ regularization. We denote $\veclambda=\{\lambda_s, \lambda_d, \lambda\}$. In the actual training, we separate the iterations of CPC data and ABSA data and input batches from the two domains alternatively. Alternative inputs ensure that the DC receives batches with different labels evenly and avoid overfitting to either domain label. A stochastic gradient descent based optimizer, Adam~\cite{adam}, is leveraged to optimize the parameters of \mn{}. Algorithm~\ref{alg:pseudocode} in Section~\ref{sec:appd_pseudocode} explains the alternative training paradigm in detail.

%% file: content/4experiments.tex
\section{Experiments}




\subsection{Experimental Settings}
\label{subsec:experiment_setting}
\paragraph{Dataset} 
CompSent-19 is the first public dataset for the CPC task released by~\citet{panchenko19}. It contains sentences with entity annotations. The ground truth is obtained by comparing the entity that appears earlier ($e_1$) in the sentence with the one that appears later ($e_2$). 
The dataset is split by convention~\cite{panchenko19,bingliu20}: 80\% for training and 20\% for testing. During training, 20\% of the training data of each label composes the development set for model selection. The detailed statistics are given in Table~\ref{tab:data_statistics}.

Three datasets of restaurants released in SemEval 2014, 2015, and 2016~\cite{semeval2014,semeval2015,semeval2016} are utilized for the ABSA task. We join their training sets and randomly sample instances into batches to optimize the auxiliary objective $\mathcal{L}_s$. The proportions of \texttt{POS}, \texttt{NEU}, and \texttt{NEG} instances are 65.8\%, 11.0\%, and 23.2\%.
\input{tables/data_statistics}

\textbf{Note.} The rigorous definition of \texttt{None} in CompSent-19 is that the sentence \textit{does not contain} a comparison between the entities rather than that entities are both preferred or disliked. Although the two definitions are not mutually exclusive, we would like to provide a clearer background of the CPC problem.

\paragraph{Imbalanced Data}
CompSent-19 is badly imbalanced (see Table~\ref{tab:data_statistics}). \texttt{None} instances dominate in the dataset. The other two labels combined only account for 27\%. This critical issue can impair the model performance. Three methods to alleviate the imbalance are tested. 
\textit{Flipping labels}: Consider the order of the entities, an original \texttt{Better} instance will become a \texttt{Worse} one and vice versa if querying $(e_2, e_1)$ instead. We interchange the $e_1$ and $e_2$ of all \texttt{Better} and \texttt{Worse} samples so that they have the same amount. 
\textit{Upsampling}: We upsample \texttt{Better} and \texttt{Worse} instances with duplication to the same amount of \texttt{None}. \textit{Weighted loss}: We upweight the underpopulated labels \texttt{Better} and \texttt{Worse} when computing the classification loss. Their effects are discussed in Section~\ref{subsec:exp_perf}.

\paragraph{Evaluation Metric}
The F1 score of each label and the micro-averaging F1 score are reported for comparison. We use F1(B), F1(W), F1(N), and micro-F1 to denote them.
The micro-F1 scores on the development set are used as the criteria to pick the best model over training epochs and the corresponding test performances are reported.

\paragraph{Reproducibility}
The implementation of \mn{} is publicly available on GitHub\footnote{\url{https://github.com/zyli93/SAECON}}. Details for reproduction are given in Section~\ref{subsec:reproducibility}.

\paragraph{Baseline Models}
Seven models experimented in~\citet{panchenko19} and the state-of-the-art ED-GAT~\cite{bingliu20} are considered for performance comparison and described in Section~\ref{sec:appd_baseline}.
\textbf{Fixed} BERT embeddings are used in our experiments same as ED-GAT for comparison fairness.

\subsection{Performances on CPC}
\label{subsec:exp_perf}
\paragraph{Comparing with Baselines}
We report the best performances of baselines and \mn{} in Table~\ref{tab:performance}. \mn{} with BERT embeddings achieves the highest F1 scores comparing with all baselines, which demonstrates the superior ability of \mn{} to accurately classify entity comparisons. The F1 scores for \texttt{None}, i.e., F1(N), are consistently the highest in all rows due to the data imbalance where \texttt{None} accounts for the largest percentage. \texttt{Worse} data is the smallest and thus is the hardest to predict precisely. This also explains why models with higher micro-F1 discussed later usually achieve larger F1(W) given that their accuracy values on the majority class (\texttt{None}) are almost identical. BERT-based models outperform GloVe-based ones, indicating the advantage of contextualized embeddings.

In later discussion, the reported performances of \mn{} and its variants are based on the BERT version. The performances of the GloVe-based \mn{} demonstrate similar trends.
\input{tables/performance}

\paragraph{Ablation Studies}
Ablation studies demonstrate the unique contribution of each part of the proposed model. Here we verify the contributions of the following modules: 
(1) The bi-directional global context extractor (BiLSTM); 
(2) The syntactic local context extractor (SGCN); 
(3) The domain adaptation modules of $\mathcal{A}$ (GRL); 
(4) The entire auxiliary sentiment analyzer, including its dependent GRL+DC ($\mathcal{A}+$GRL for short). The results are presented in Table~\ref{tab:ablation}. 

Four points worth noting. 
Firstly, the \mn{} with all modules achieves the best performance on three out of four metrics, demonstrating the effectiveness of all modules (\mn{} vs. the rest); 
Secondly, the synergy of $\mathcal{A}$ and GRL improves the performance ($-(\mathcal{A}+$GRL$)$ vs. \mn{}) whereas the $\mathcal{A}$ without domain adaptation hurts the classification accuracy instead ($-$GRL vs. \mn{}), which indicates that the auxiliary sentiment analyzer is beneficial to CPC accuracy only with the assistance of GRL+DC modules;
Thirdly, removing the global context causes the largest performance deterioration ($-$BiLSTM vs. \mn{}), showing the significance of long-term information. This observation is consistent with the findings of~\citet{bingliu20} in which eight to ten stacking GAT layers are used for global feature learning; 
Finally, the performances also drop after removing the SGCN ($-$SGCN vs. \mn{}) but the drop is less than removing the BiLSTM. Therefore, local context plays a less important role than the global context ($-$SGCN vs. $-$BiLSTM).
\input{tables/ablation}

\paragraph{Hyperparameter Searching}
We demonstrate the influences of several key hyperparameters. 
Such hyperparameters include the initial learning rate (LR, $\eta$), feature dimensions ($\vecd=\{d_g,d_l, d_s\}$), regularization weight $\lambda$, and the configurations of SGCN such as directionality, gating, and layer numbers.

For LR, $\vecd$, and $\lambda$ in Figures~\ref{fig:lr},~\ref{fig:dim}, and~\ref{fig:lambda}, we can observe a single peak for F1(W) (green curves) and fluctuating F1 scores for other labels and the micro-F1 (blue curves). In addition, the peaks of micro-F1 occur at the same positions of F1(W). This indicates that the performance on \texttt{Worse} is the \textbf{most} influential factor to the micro-F1. These observations help us locate the optimal settings and also show the strong learning stability of \mn{}.

Figure~\ref{fig:sgcn} focuses on the effect of SGCN layer numbers. We observe clear oscillations on F1(W) and find the best scores at two layers. More layers of GCN result in oversmoothing~\cite{gcn} and hugely downgrade the accuracy, which is eased but not entirely fixed by the gating mechanism. Therefore, the performances slightly drop on larger layer numbers. 

Table~\ref{tab:hyperparam_dg} shows the impact of directionality and gating. Turning off either the directionality or the gating mechanism (``\xmark{}\cmark{}'' or ``\cmark{}\xmark{}'') leads to degraded F1 scores. SGCN without modifications (``\xmark{}\xmark{}'') drops to the poorest micro-F1 and F1(W). Although its F1(N) is the highest, we hardly consider it a good sign. Overall, the benefits of the directionality and gating are verified.

\input{tables/hyperparam_search}
\paragraph{Alleviating Data Imbalance}
The label imbalance severely impairs the model performance, especially on the most underpopulated label \texttt{Worse}. 
The aforementioned imbalance alleviation methods are tested in Table~\ref{tab:data_aug}. The \textit{Original} (OR) row is a control experiment using the raw CompSent-19 without any weighting or augmentation. 

The optimal solution is the weighted loss (WL vs. the rest). One interesting observation is that data augmentation such as flipping labels and upsampling cannot provide a performance gain (OR vs. FL and OR vs. UP). Weighted loss performs a bit worse on F1(N) but consistently better on the other metrics, especially on \texttt{Worse}, indicating that it effectively alleviates the imbalance issue. In practice, static weights found via grid search are assigned to different labels when computing the cross entropy loss. We leave the exploration of dynamic weighting methods such as the Focal Loss~\cite{lin2017focal} for future work. 
\input{tables/data_aug}
\input{tables/case_study}

\paragraph{Alternative Training}
One novelty of \mn{} is the alternative training that allows the sentiment analyzer to learn both tasks across domains. Here we analyze the impacts of different batch ratios (BR) and different domain shift handling methods during the training. BR controls the number of ratio of batches of the two alternative tasks in each training cycle. For example, a BR of $2:3$ sends 2 CPC batches followed by 3 ABSA batches in each iteration. 

Figure~\ref{fig:br_vs_time} presents the entire training time for ten epochs with different BR. A larger BR takes shorter time. For example, a BR of 1:1 (the leftmost bar) takes a shorter time than 1:5 (the yellow bar). Figure~\ref{fig:br_vs_f1} presents the micro-F1 scores for different BR. We observe two points: (1) The reported performances differ slightly; (2) Generally, the performance is better when the CPC batches are less than ABSA ones. Overall, the hyperparameter selection tries to find a ``sweet spot'' for effectiveness and efficiency, which points to the BR of 1:1.
\input{tables/alt_training}

Figure~\ref{fig:domainshift} depicts the performance comparisons of \mn{} (green bars), \mn{}$-$GRL (the ``$-$GRL'' in Figure~\ref{tab:ablation}, orange bars), and \mn{} with pretrained and fixed parameters of $\mathcal{A}$ (the option (1) mentioned in Section~\ref{subsec:knowledge_transfer}, blue bars). They represent different levels of domain shift mitigation: The pretrained and fixed $\mathcal{A}$ \textit{does NOT} handle the domain shift at all; The variant $-$GRL only attempts to implicitly handle the shift by alternative training with different tasks to converge in the middle although the domain difference can be harmful to both objectives; \mn{}, instead, explicitly uses GRL+DC to mitigate the domain shift between $\mathcal{D}_s$ and $\mathcal{D}_c$ during training. 

As a result, \mn{} achieves the best performance especially on F1(W), $-$GRL gets the second, and the ``option (1)'' gets the worst. These demonstrate that (1) the alternative training (blue vs. green) for an effective domain adaptation is necessary and (2) there exists a positive correlation between the level of domain shift mitigation and the model performance, especially on F1(W) and F1(B). A better domain adaptation produces higher F1 scores in the scenarios where datasets in the domain of interest, i.e., CPC, is unavailable. 
\input{tables/domainshift}

\subsection{Case Study}
\label{subsec:case_study}
In this section, we qualitatively exemplify the contribution of the sentiment analyzer $\mathcal{A}$.
Table~\ref{tab:case_study} reports four example sentences from the test set of CompSent-19.
The entities $e_1$ and $e_2$ are highlighted together with the corresponding sentiment predicted by $\mathcal{A}$. The column ``Label'' shows the ground truth of CPC. The ``$\Delta$'' column computes the sentiment distances between the entities. 
We assign $+1$, $0$, and $-1$ to sentiment polarities of \texttt{POS}, \texttt{NEU}, and \texttt{NEG}, respectively.
$\Delta$ is computed by the sentiment polarity of $e_1$ minus that of $e_2$. Therefore, a positive distance suggests that $e_1$ receives a more positive sentiment from $\mathcal{A}$ than $e_2$ and vice versa.
In \textbf{S1}, sentiments to \textit{Ethernet} and \textit{USB} are predicted positive and negative, respectively, which can correctly imply the comparative label as \texttt{Better}. \textbf{S2} is a \texttt{Worse} sentence with \textit{Bash} predicted negative, \textit{Python} predicted neutral, and a resultant negative sentiment distance $-1$.
For \textbf{S3} and \textbf{S4}, the entities are assigned the same polarities. Therefore, the sentiment distances are both zeros. We can easily tell that preference comparisons do not exist, which is consistent with the ground truth labels. Due to the limited space, more interesting case studies are presented in Section~\ref{sec:appd_case_study}.

%% file: tables/data_statistics.tex
\begin{table}[!h]
    \centering
    \resizebox{\linewidth}{!}{
    \setlength\tabcolsep{2pt}
    \begin{tabular}{c|cccc}
    \hline  \hline
    \textbf{Dataset} & \texttt{Better} &\texttt{Worse} & \texttt{None} &\textbf{Total} \\ \hline
     Train & 872 (19\%) & 379 (8\%) & 3,355 (73\%) & 4,606 \\
     Development & 219 (19\%) & 95 (8\%)  & 839 (73\%) & 1,153 \\ 
     Test & 273 (19\%)  & 119 (8\%) & 1,048 (73\%) & 1,440\\ \hline
     Total & 1,346 (19\%)  & 593 (8\%) &  5,242 (73\%) & 7,199 \\ \hline \hline
     Flipping labels & 1,251 (21\%) & 1,251 (21\%) & 3,355 (58\%) & 5,857 \\
     Upsampling & 3,355 (33\%) & 3,355 (33\%) & 3,355 (33\%) & 10,065 \\ \hline \hline
    \end{tabular}
    }
    \caption{Statistics of CompSent-19. The rows of \textit{Flipping labels} and \textit{Upsampling} show the numbers of the augmented datasets to mitigate label imbalance.}
    \label{tab:data_statistics}
\end{table}

%% file: tables/performance.tex
\begin{table}[!h]
    \centering
    \resizebox{0.9\linewidth}{!}{
    \setlength\tabcolsep{2pt}
    \begin{tabular}{c|cccc}
    \hline  
    \textbf{Model} & \textbf{Micro.} &\textbf{F1(B)} & \textbf{F1(W)} &\textbf{F1(N)} \\ \hline
    Majority & 68.95 & 0.0    & 0.0    & 81.62 \\
    SE-Lin   & 79.31 & 62.71  & 37.61  & 88.42 \\
    SE-XGB   & 85.00 & \underline{75.00}  & 43.00  & 92.00 \\ 
    SVM-Tree & 68.12 & 53.35  & 13.90  & 78.13 \\ \hline
    BERT-CLS & 83.12 & 69.62  & \underline{50.37}  & 89.84 \\
    AvgWE-G  & 76.32 & 48.28  & 20.12  & 86.34 \\
    AvgWE-B  & 77.64 & 53.94  & 26.88  & 87.47 \\ \hline
    ED-GAT-G & 82.73 & 70.23 & 43.30   & 89.84 \\
    ED-GAT-B & \underline{85.42} & 71.65 & 47.29   & \underline{92.34} \\ \hline
    \rowcolor[gray]{.92}
     \mn{}-G & 83.78 & 71.06 & 45.90 & 91.05 \\ 
    \rowcolor[gray]{.92}
     \mn{}-B & \textbf{86.74} & \textbf{77.10} & \textbf{54.08} & \textbf{92.64} \\
     \hline 
    \end{tabular}
    }
    \caption{Performance comparisons between the proposed model and baselines on F1 scores (\%). ``-B'' and ``-G'' denote different versions of the model using BERT~\cite{bert} and GloVe~\cite{pennington2014glove} as the input embeddings, respectively. All reported improvements over the best baselines are statistically significant with $p$-value $<$ 0.01.}
    \label{tab:performance}
\end{table}

%% file: tables/ablation.tex
\begin{table}[!h]
    \centering
    \resizebox{0.95\linewidth}{!}{
    \setlength\tabcolsep{3pt}
    \begin{tabular}{c|cccc} \hline 
    \textbf{Variants} & \textbf{Micro.} &\textbf{F1(B)} & \textbf{F1(W)} &\textbf{F1(N)} \\ \hline
    \rowcolor[gray]{.92}
    \mn{} &  \textbf{86.74} & \textbf{77.10} & \textbf{54.08} & 92.64 \\ \hline 
    $-$BiLSTM             & 85.21 & 72.94 & 43.86 & 92.63 \\
    $-$SGCN                & 86.53 & 76.22 & 51.38 & 92.24 \\ \hline
    $-$GRL                & 86.53 & 76.16 & 49.77 & \textbf{92.93} \\ 
    $-(\mathcal{A}+$GRL$)$ & 85.97 & 74.82 & 52.44 & 92.45 \\ \hline
    \end{tabular}
    }
    \caption{Ablation studies on F1 scores (\%) between \mn{} and its variants with modules disabled. ``$-$X'' denotes a variant without module X. Removing $\mathcal{A}$ will also remove GRL+DC.}
    \label{tab:ablation}
\end{table}

%% file: tables/hyperparam_search.tex
\begin{figure}[t!]
    \centering
    \begin{subfigure}[t]{0.5\linewidth}
    \captionsetup{width=0.95\linewidth}
        \centering
        \includegraphics[width=0.98\linewidth]{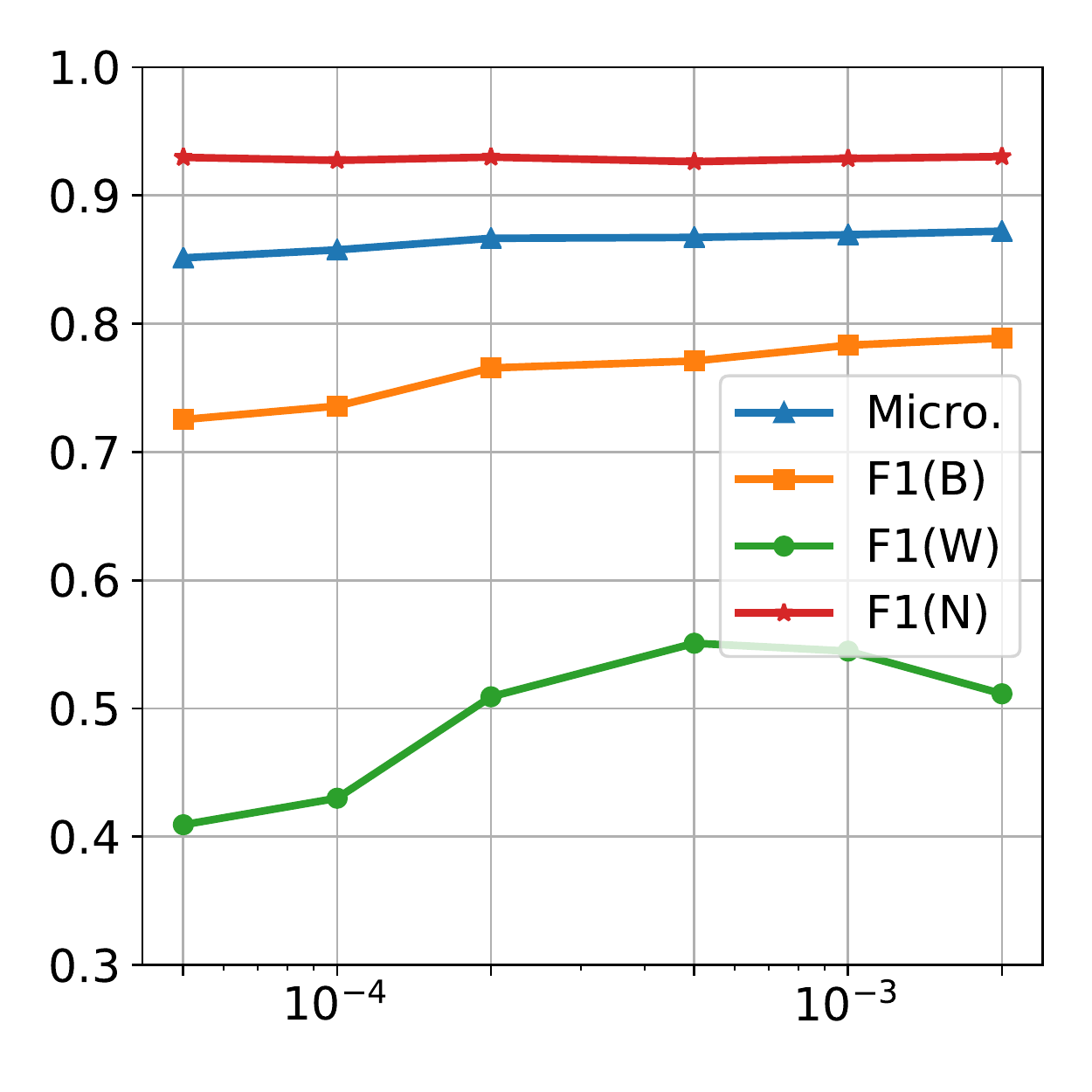}
        \caption{F1 vs. Learning rate ($\eta$)}
        \label{fig:lr}
    \end{subfigure}%
    ~ 
    \begin{subfigure}[t]{0.5\linewidth}
    \captionsetup{width=0.95\linewidth}
        \centering
        \includegraphics[width=0.98\linewidth]{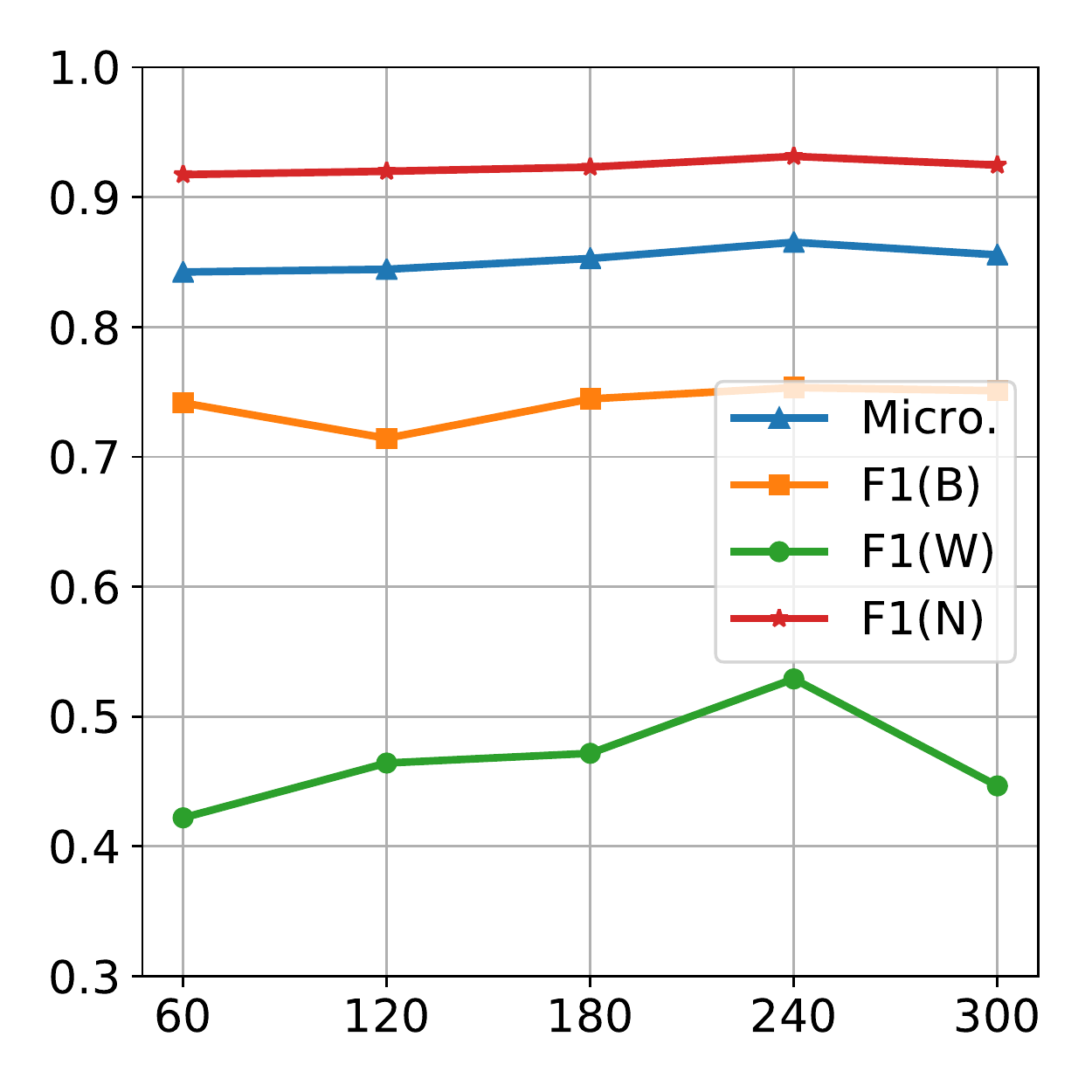}
        \caption{F1 vs. Feature dims. ($\vecd$)}
        \label{fig:dim}
    \end{subfigure}
    \begin{subfigure}[t]{0.5\linewidth}
    \captionsetup{width=0.95\linewidth}
        \centering
        \includegraphics[width=0.98\linewidth]{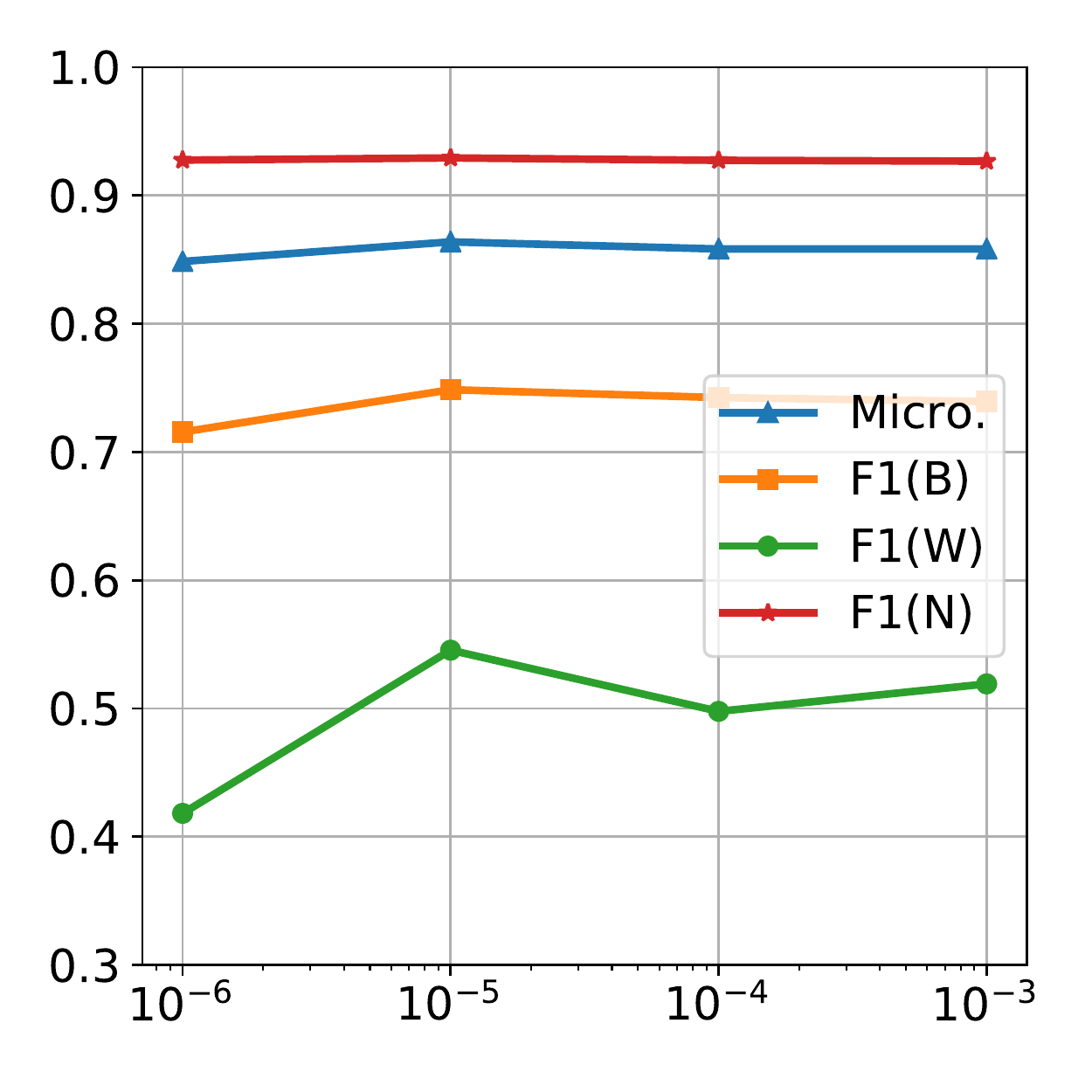}
        \caption{F1 vs. Reg. weight ($\lambda$)}
        \label{fig:lambda}
    \end{subfigure}%
    ~ 
    \begin{subfigure}[t]{0.5\linewidth}
    \captionsetup{width=0.95\linewidth}
        \centering
        \includegraphics[width=0.98\linewidth]{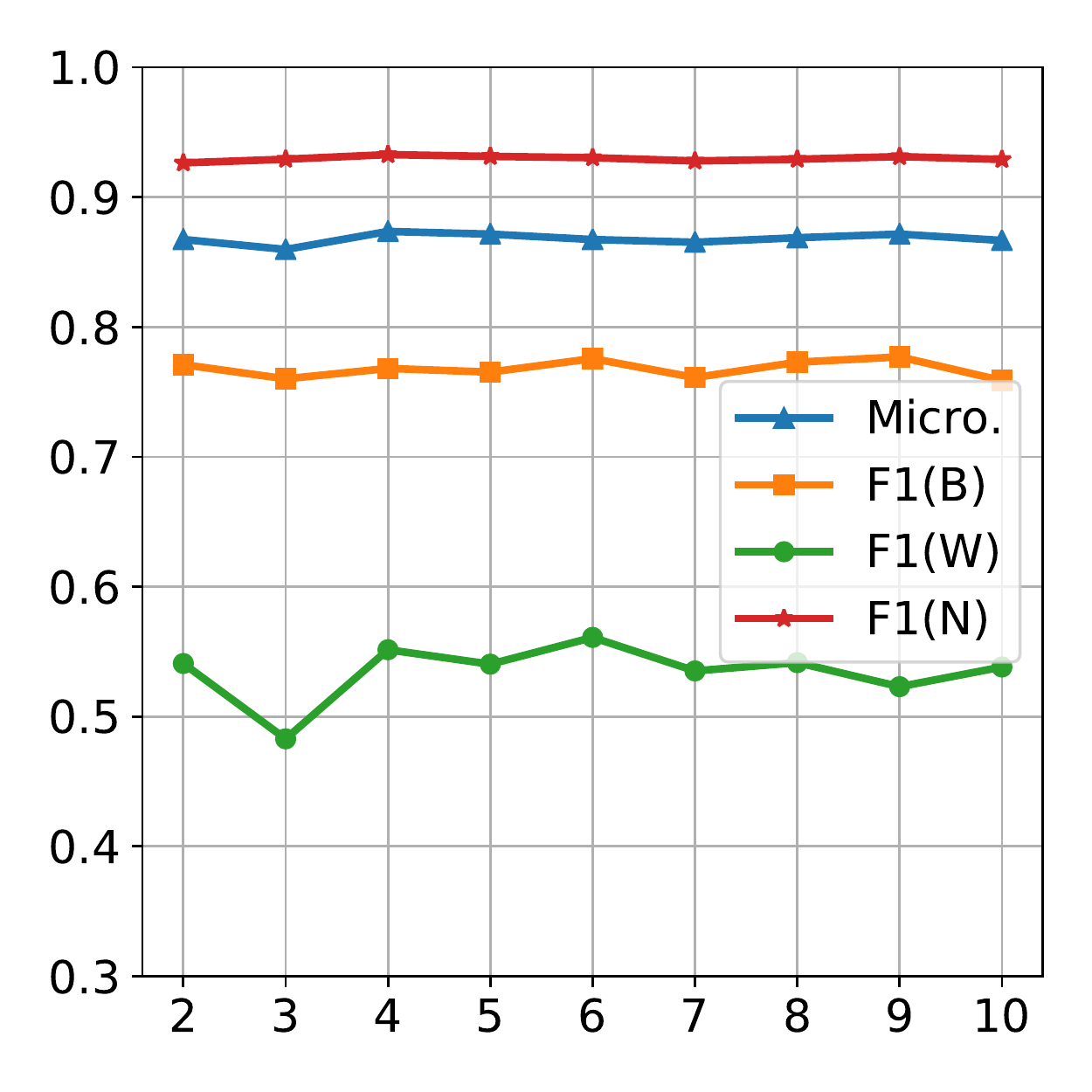}
        \caption{F1 vs. SGCN layers}
        \label{fig:sgcn}
    \end{subfigure}
    \caption{Searching and sensitivity for four key hyperparameters of \mn{} in F1 scores.}
    \label{fig:hyperparam}
\end{figure}
\begin{table}[!h]
    \centering
    \resizebox{0.95\linewidth}{!}{
       \setlength\tabcolsep{2pt}
    \begin{tabular}{cc|cccc} \hline
    \textbf{Directed} & \textbf{Gating}  & \textbf{Micro.} &\textbf{F1(B)} & \textbf{F1(W)} &\textbf{F1(N)} \\ \hline
    \rowcolor[gray]{.92}
     \cmark & \cmark &  \textbf{86.74} & \textbf{77.10} & \textbf{54.08} & 92.64 \\  \hline
     \xmark & \cmark & 86.18 & 75.72 & 49.78 &  92.40 \\
     \cmark & \xmark & 85.35 & 74.03 & 43.27 &  92.34 \\ 
     \xmark & \xmark & 85.35 & 73.39 & 35.78 &  \textbf{93.04} \\ \hline
    \end{tabular}
    }
    \caption{Searching and sensitivity for the directionality and gating of SGCN by F1 scores (\%).}
    \label{tab:hyperparam_dg}
\end{table}

%% file: tables/data_aug.tex
\begin{table}[!h]
    \centering
    \resizebox{0.95\linewidth}{!}{
    \setlength\tabcolsep{3pt}
    \begin{tabular}{c|cccc}
    \hline  
    \textbf{Methods} & \textbf{Micro.} &\textbf{F1(B)} & \textbf{F1(W)} &\textbf{F1(N)} \\ \hline
    \rowcolor[gray]{.92}
     Weighted loss (WL)  & \textbf{86.74} & \textbf{77.10} & \textbf{54.08} & 92.64 \\ \hline
     Original (OR)       & 85.97 & 73.80 & 46.15 & 92.90 \\
     Flipping labels (FL)   & 84.93 & 73.07 & 42.45 & 91.99 \\
     Upsampling (UP)   & 85.83 & 73.11 & 46.36 & \textbf{92.95} \\
     \hline 
    \end{tabular}
    }
    \caption{Performance analysis with F1 scores (\%) for different methods to mitigate data imbalance.}
    \label{tab:data_aug}
\end{table}

%% file: tables/case_study.tex
\newcommand{\entA}[2]{[\textbf{\textcolor[HTML]{2E9E2E}{#1}}:\texttt{{#2}}]} 
\newcommand{\entB}[2]{[\textbf{\textcolor[HTML]{CC0000}{#1}}:\texttt{{#2}}]} 

\newcommand{\sentBetterOne}{\small{\textbf{S1}: This is all done via the gigabit \entA{Ethernet}{POS} interface, rather than the much slower \entB{USB}{NEG} interface.}}
\newcommand{\sentBetterTwo}{\small{\textbf{S1}: \entA{Ruby}{NEU} wasn't designed to ``exemplify best practices'', it was to be a better \entB{Perl}{NEG}.}}

\newcommand{\sentWorseOne}{\small{\textbf{S2}: Also, \entA{Bash}{NEG} may not be the best language to do arithmetic heavy operations in something like \entB{Python}{NEU} might be a better choice.}}
\newcommand{\sentWorseTwo}{\small{\textbf{S2}: And from my experience the ticks are much worse in \entA{Mid Missouri}{NEG} than they are in \entB{South Georgia}{POS} which is much warmer year round.}}

\newcommand{\sentNoneOne}{\small{\textbf{S3}: It shows how \entA{JavaScript}{POS} and \entB{PHP}{POS} can be used in tandem to make a user's experience faster and more pleasant.}}
\newcommand{\sentNoneTwo}{\small{\textbf{S4}: He broke his hand against \entA{Georgia Tech}{NEU} and made it worse playing against \entB{Virginia Tech}{NEU}.}}
\newcommand{\sentNoneThree}{\small{\textbf{S3}: As an industry rule, \entA{hockey}{NEG} and \entB{basketball}{NEG} sell comparatively poorly everywhere.}}
\newcommand{\sentNoneFour}{\small{\textbf{S4}: \entA{Milk}{NEG}, \entB{juice}{NEG} and soda make it ten times worse.}}

\begin{table*}[!h]
    \centering
    \resizebox{0.95\linewidth}{!}{
    \setlength\tabcolsep{3pt}
    \begin{tabular}{m{14cm}|c|c} \hline \hline
    \centering{CPC sentences with sentiment predictions by $\mathcal{A}$} & Label & $\Delta$ \\ \hline
    \sentBetterOne & Better & $+2$  \\ \hline
    \sentWorseOne  & Worse  & $-1$ \\ \hline
    \sentNoneOne   & None   & $0$  \\ \hline
    \sentNoneTwo   & None   & $0$  \\ \hline \hline
    \end{tabular}
    }
    \caption{Case studies for the effect of the sentiment analyzer $\mathcal{A}$ (see Section~\ref{subsec:case_study} for details).}
    \label{tab:case_study}
\end{table*}

%% file: tables/alt_training.tex
\begin{figure}[t!]
    \centering
    \begin{subfigure}[t]{0.5\linewidth}
    \captionsetup{width=\linewidth}
        \centering
        \includegraphics[width=\linewidth]{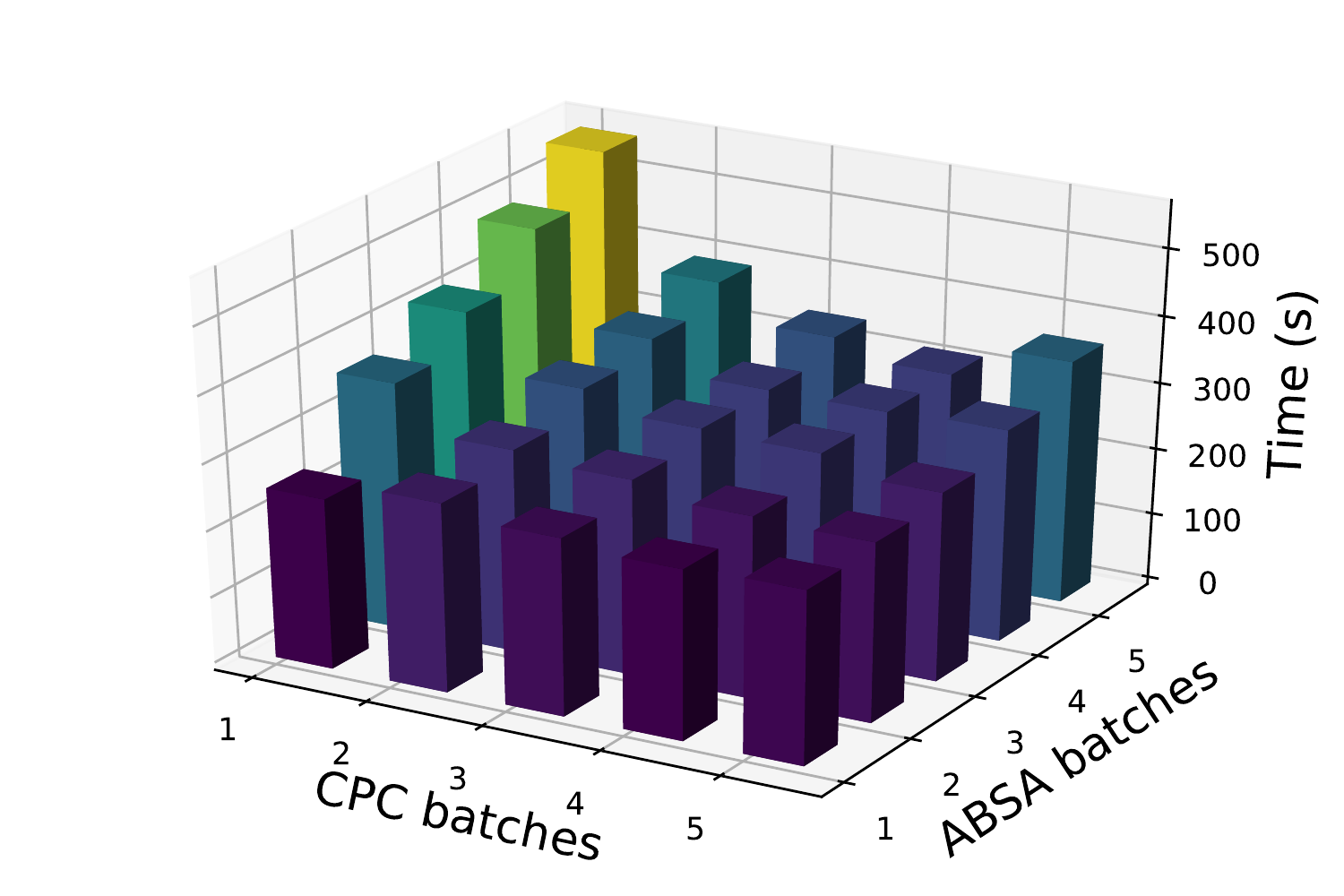}
        \caption{BR vs. Training time (s)}
        \label{fig:br_vs_time}
    \end{subfigure}%
    ~ 
    \begin{subfigure}[t]{0.5\linewidth}
    \captionsetup{width=\linewidth}
        \centering
        \includegraphics[width=\linewidth]{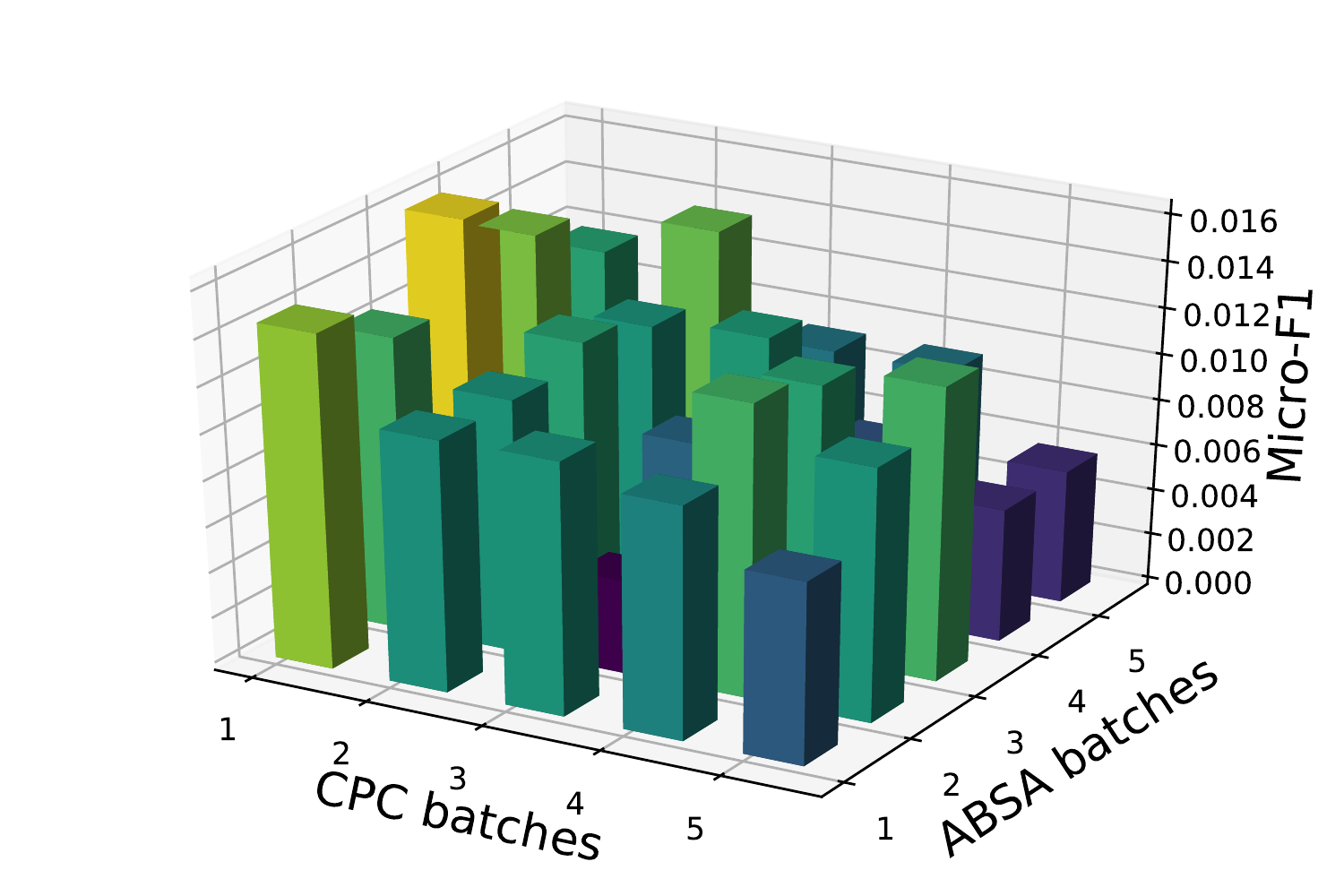}
        \caption{BR vs. Micro-F1}
        \label{fig:br_vs_f1}
    \end{subfigure}
    \caption{Analyses for batch ration (BR). The values of micro-F1 are actual numbers minus 0.85 for the convenience of visualization.}
    \label{fig:alternative}
\end{figure}

%% file: tables/domainshift.tex
\begin{figure}[!h]
    \centering
    \includegraphics[width=0.95\linewidth]{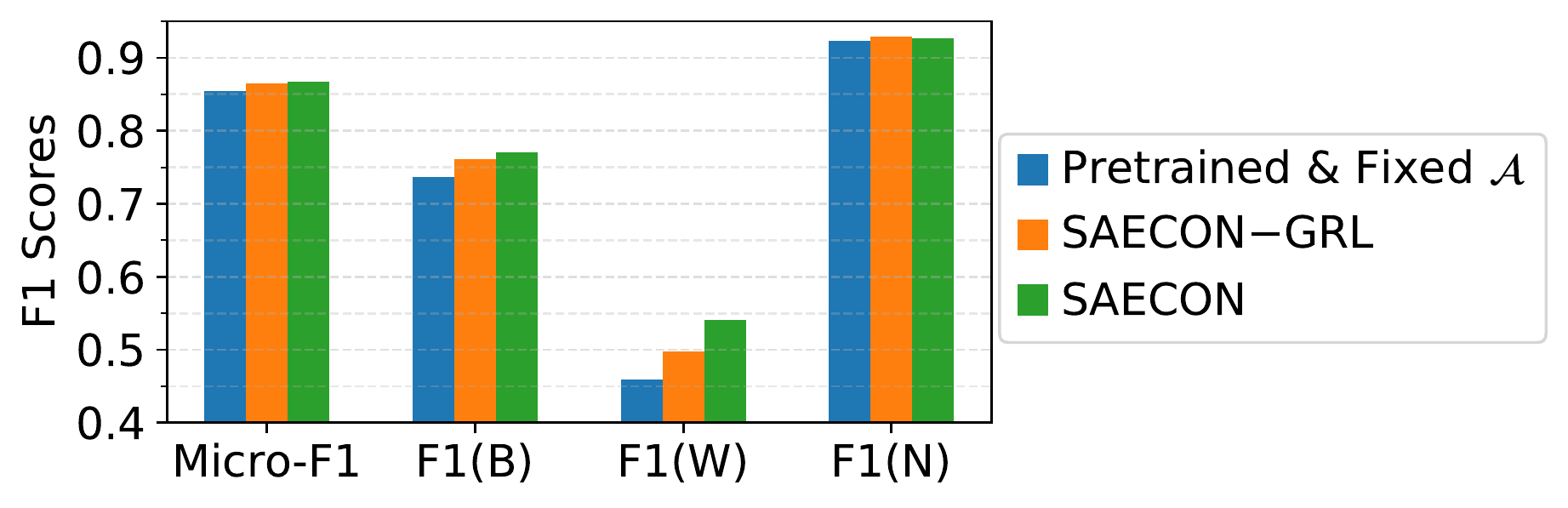}\vspace{-0.5em}
    \caption{Visualization of the domain shift mitigation.}
    \label{fig:domainshift}\vspace{-1em}
\end{figure}

%% file: content/5conclusion.tex
\section{Conclusion}
\label{sec:conclusion}
This paper proposes \mn{}, a CPC model that incorporates a sentiment analyzer to transfer knowledge from ABSA corpora. Specifically, \mn{} utilizes a BiLSTM to learn global comparative features, a syntactic GCN to learn local syntactic information, and a domain adaptive auxiliary sentiment analyzer that jointly learns from ABSA corpora and CPC corpora for a smooth knowledge transfer. An alternative joint training scheme enables the efficient and effective information transfer. Qualitative and quantitative experiments verified the superior performance of \mn{}. For future work, we will focus on a deeper understanding of CPC data augmentation and an exploration of weighting loss methods for data imbalance.

%% file: content/8acknowledgement.tex
\section*{Acknowledgement}
We would like to thank the reviewers for their helpful comments. 
The work was partially supported by NSF DGE-1829071 and NSF IIS-2106859.

%% file: content/7broaderimpact.tex
\section*{Broader Impact Statement}

This section states the broader impact of the proposed CPC model. Our model is designed specifically for the comparative classification scenarios in NLP. Users can use our model to detect whether a comparison exists between two entities of interest within the sentences of a particular sentential corpus. For example, a recommender system equipped with our model can tell whether a product is compared with a competitor and, further, which is preferred. In addition, Review-based platforms can utilize our model to decide which items are widely welcomed and which are not.

Our model, \mn{}, is tested with the CompSent-19 dataset which has been anonymized during the process of collection and annotation. For the auxiliary ABSA task, we also use an anonymized public dataset from SemEval 2014 to 2016. Therefore, our model will not cause potential leakages of user identity and privacy.
We would like to call for attention to CPC as it is a relatively new task in NLP research. 

%% file: content/6appendix.tex
\section{Supplementary Materials}
\label{sec:appendix}

This section contains the supplementary materials for \textit{\myTitle{}}. Here we provide additional supporting information in four aspects, including additional description for the training, the reproducibility details of \mn{}, brief introductions of baselines, and additional case studies.

\subsection{Pseudocode of \mn{}}
\label{sec:appd_pseudocode}
In this section, we show the pseudocode of the training of \mn{} to provide a comprehensive picture of the alternative training paradigm.
\input{tables/pseudocode}

\subsection{Reproducibility}
\label{subsec:reproducibility}
In this section, we provide the instructions to reproduce \mn{} and the default hyperparameter settings to generate the performances reported in Section~\ref{subsec:exp_perf}. 

\subsubsection{Implementation of \mn{}}
The proposed \mn{} is implemented in Python (3.6.8) with PyTorch (1.5.0) and run with a single 16GB Nvidia V100 GPU. 
The source code of \mn{} is publicly available on GitHub\footnote{The source code will be publicly available if the paper is accepted. A copy of anonymized source code is submitted for review.} and comprehensive instructions on how to reproduce our model are also provided. 

The implementation of SGCN is based on PyTorch Geometric\footnote{\url{https://github.com/rusty1s/pytorch_geometric}}. 
The implementation of our sentiment analyzer $\mathcal{A}$ is adapted from the official source code of LCF-ASC~\cite{phan-ogunbona-2020-modelling}\footnote{\url{https://github.com/HieuPhan33/LCFS-BERT}}. 
The dependency parser used in \mn{} is from spaCy~\footnote{\url{https://spacy.io/}}.
The pretrained embedding vectors of GloVe are downloaded from the office site\footnote{\url{https://nlp.stanford.edu/projects/glove/}}. 
The pretrained BERT model is obtained from the Hugging Face model repository\footnote{\url{https://huggingface.co/bert-base-uncased}}. 
The implementation of the gradient reversal package is available on GitHub\footnote{\url{https://github.com/janfreyberg/pytorch-revgrad}}.
We would like to appreciate the authors of these packages for their precious contributions.

\subsubsection{Default Hyperparameters}
The default hyperparameter settings for the results reported in Section~\ref{subsec:exp_perf} are given in Table~\ref{tab:hyperparam}
\input{tables/default_hyperparam}

\input{tables/appd_case_study}
\subsubsection{Reproduction of ED-GAT}
We briefly introduce the reproduction of the state-of-the-art baseline, ED-GAT, in both GloVe and BERT versions. We implement ED-GAT with the same software packages as \mn{} such as PyTorch-Geometric, spaCy, and PyTorch, and run it within the same machine environment. The parameters all follow the original ED-GAT setting~\cite{bingliu20} except the dimension of GloVe. It is set to 300 in the original paper but 100 in our experiments for the fairness of comparison. The number of layers is select as 8 and the hidden size is set to 300 for each layer with 6 attention heads. We trained the model for 15 epochs with Adam optimizer with a batch size of 32.

\subsection{Baseline models}
\label{sec:appd_baseline}
We briefly introduce the compared models in Section~\ref{subsec:exp_perf}.

\indent\textbf{Majority-Class} A simple model which chooses the majority label in the training set as the prediction of each test instance.

\indent\textbf{SE} Sentence Embedding encodes the sentences into low-dimensional sentence representations using pretrained language encoders~\cite{conneau17supervised,dowman15large} and then feeds them into a classifier for comparative preference prediction. \textbf{SE} has two versions~\cite{panchenko19} with different classifiers, namely \textbf{SE-Lin} with a linear classifier and \textbf{SE-XGB} with an XGBoost classifier.

\indent\textbf{SVM-Tree} This method~\cite{svmtree} applies convolutional kernel methods to CSI task. We follow the experimental settings of~\citep{bingliu20}.

\indent\textbf{AvgWE} A word embedding-based method that averages the word embeddings of the sentence as the sentence representation and then feeds this representation into a classifier. The input embeddings have several options, such as GloVe~\cite{pennington2014glove} and BERT~\cite{bert}. These variants are denoted by \textbf{AvgWE-G} and \textbf{AvgWE-B} separately.

\indent\textbf{BERT-CLS} Using the representation of the token ``[CLS]'' generated by BERT~\cite{bert} as the sentence embedding and a linear classifier to conduct comparative preference classification.

\indent\textbf{ED-GAT} Entity-aware Dependency-based Graph Attention Network~\cite{bingliu20} is the first entity-aware model that analyzes the entity relations via the dependency graph and multi-layer graph attention layer. 

\subsection{Additional Case Studies}
\label{sec:appd_case_study}
In this section, we present four supplementary examples for case study in Table~\ref{tab:appd_case_study} which have different sentiments compared with their counterparts in Table~\ref{tab:case_study}. \textbf{S1} shows a \texttt{NEU} versus \texttt{NEG} comparison which results in a sentiment distance $+1$ and a CPC prediction \texttt{Better}. ``Ruby'' is not praised in this sentence so it has \texttt{NEU}. But ``Perl'' is assigned a negative emotion through a simple inference. \textbf{S2} shows a stronger contrast between the entities. ``Mid Missouri'' is said ``much worse'' while the ``South Georgia'' is ``much warmer'', which clearly indicates the sentiments and the comparative classification results.

\textbf{S3} and \textbf{S4} are two sentences both with two parallel negative entities. The sport equipment in \textbf{S3} is sold ``poorly'' and the drinks in \textbf{S4} are ``ten times worse'' both indicating negative sentiments. Therefore, the labels are \texttt{None}. 

%% file: tables/pseudocode.tex
\newcommand\mycommfont[1]{\footnotesize\ttfamily\textcolor{blue}{#1}}
\SetArgSty{textnormal}
\SetCommentSty{mycommfont}
\SetKw{And}{and}
\SetKw{Or}{or}
\SetKw{Ret}{return}
\SetKwInput{KwInput}{Input}                
\SetKwInput{KwOutput}{Output}              

\begin{algorithm}[h]
\DontPrintSemicolon
  \KwInput{Loss weights $\veclambda$; Learning rate $\eta$}
  \KwData{$D_c$ and $D_s$.}
  \While{not converge}{
    $\{s,E\}$, \textit{task} $\xleftarrow{}$ getAltSample($D_c$, $D_s$) \;
    $\matS_0 \xleftarrow{}$ TextEncode($s$) \;
    $G_s \xleftarrow{}$ DepParse($s$) \;
    \eIf{\textit{task} is \texttt{CPC}}{
        $\{\vech_{g,i}\}, \{\vech_{l,i}\}$ $(i=1,2) \xleftarrow{}$ methods in Section~\ref{subsec:cpc_features}. \;
        $\vech_{s,1},\vech_{s,2} \xleftarrow{}$ $\mathcal{A}(\matS_0,G_s,E)$ \;
        $\mathcal{L}_c,\mathcal{L}_d \xleftarrow{}$ methods in Section~\ref{subsec:obj_optim}\;
        optimize($\{\mathcal{L}_c,\lambda_d\mathcal{L}_d, \lambda\text{reg}(L2)\}, \eta$) \;
        }{\tcp{sentiment analysis}
        $\vech_{s} \xleftarrow{}$ $\mathcal{A}(\matS_0, G_s, E)$ \;
        $\mathcal{L}_s,\mathcal{L}_d \xleftarrow{}$ methods in Section~\ref{subsec:obj_optim}\;
        optimize($\{\lambda_s\mathcal{L}_s,\lambda_d\mathcal{L}_d, \lambda\text{reg}(L2)\}, \eta$)
        }
    }
\caption{Optimization of \mn{} with two alternative tasks}
\label{alg:pseudocode}
\end{algorithm}

%% file: tables/default_hyperparam.tex
\begin{table}[!h]
    \centering
    \resizebox{0.99\linewidth}{!}{%
    \begin{tabular}{c|c}
    \hline \hline
    Hyperparameter    & Setting\\ \hline
    GloVe embeddings    &  pretrained, 100 dims ($d_0$)\\
    BERT version       & \texttt{bert-base-uncased} \\
    BERT num.       & 12 heads, 12 layers, 768 dims ($d_0$)\\
    BERT param.       & pretrained by Hugging Face  \\
    Dependency parser & spaCy, pretrained model \\ \hline
    Batch config. & size $=16$, batch ratio $=1:1$\\
    Init. LR ($\eta$) & $5\times 10 ^{-4}$ \\
    CPC loss weight & $2:4:1$ (B:W:N) \\ 
    $\veclambda$ $(\{\lambda,\lambda_s, \lambda_d$\}) & $\{1\times 10^{-4}, 1,1\}$ \\
    Activation & ReLU ($f(x)=\max(0,x)$)\\
    $\vecd$ $(\{d_g,d_l,d_s\})$ & $\{240,240,240\}$ \\
    Optimizer & Adam $(\beta_1=0.9, \beta_2=0.999)$\\
    LR scheduler & \texttt{StepLR} (steps $=3$, $\gamma=0.8$)\\ 
    GRL config. &  $\alpha=1.0$ (Default setting)\\
    SGCN num. & 2 layers (768$\xrightarrow{}$256$\xrightarrow{}$240) \\
    SGCN arch. & Directed, Gated \\
    Data aug. & Off (weighted loss only)\\  \hline \hline
    \end{tabular}
    }
    \caption{Default hyperparameter settings. ``LR'' is short for learning rate. ``num.'' shows the numerical configurations and ``arch.'' shows the architectural configurations. ``aug.'' is short for augmentation.}
    \label{tab:hyperparam}
\end{table}

%% file: tables/appd_case_study.tex
\begin{table*}[!h]
    \centering
    \resizebox{\linewidth}{!}{
    \setlength\tabcolsep{3pt}
    \begin{tabular}{m{14cm}|c|c} \hline \hline
    \centering{Supplementary CPC sentences with sentiment predictions by $\mathcal{A}$} & Label & $\Delta$ \\ \hline
    \sentBetterTwo & Better & $+1$  \\ \hline
    \sentWorseTwo  & Worse  & $-2$ \\ \hline
    \sentNoneThree  & None    & $0$ \\ \hline
    \sentNoneFour & None   & $0$  \\ \hline \hline
    \end{tabular}
    }
    \caption{Additional case studies for the effect of sentiment analyzer $\mathcal{A}$ (see Section~\ref{sec:appd_case_study} for details).}
    \label{tab:appd_case_study}
\end{table*}